\documentclass{article}

\usepackage[dblblindworkshop, preprint]{neurips_2026}

\workshoptitle{Foundation Models for the Brain and Body}

\usepackage[utf8]{inputenc} 
\usepackage[T1]{fontenc}    
\usepackage{hyperref}       
\usepackage{url}            
\usepackage{booktabs}       
\usepackage{amsfonts}       
\usepackage{nicefrac}       
\usepackage{microtype}      
\usepackage{xcolor}         
\usepackage{amsmath}
\usepackage{graphicx}

\title{Do sEMG Foundation Models Improve Cross-Subject Generalisation?}

\author{%
  William ~Raftery \\
  Imperial College London\\
  \texttt{w.raftery24@imperial.ac.uk} \\
  \And
  Dario ~Farina \\
  Imperial College London\\
}

\begin{document}

\maketitle

\begin{abstract}

sEMG Foundation Models aim to learn general purpose representations over diverse subjects and recording conditions that boost performance on a variety of downstream tasks. However, existing evaluations make it difficult to discern whether pretraining alone is sufficient to combat the cross-subject overfitting which makes machine learning on sEMG so challenging. We conduct a systematic evaluation of three sEMG foundation models across three gesture classification datasets under both intra-session and inter-subject splits, using three fine-tuning strategies designed to isolate the contribution of pretraining. We find that pretraining boosts performance when train and test samples are drawn from the same recording session, but this benefit largely disappears under inter-subject evaluation. In low subject count datasets, pretraining provides little to no benefit and classical ML baselines match or exceed foundation model performance.

Code is available at: \url{https://github.com/willraf/emg-fm-bench}

\end{abstract}

\section{Introduction}

Through non-invasive measurement of neuromuscular activity, surface electromyography (sEMG) encodes information spanning motor output, neural drive and neuromuscular physiology \cite{merlettiSurfaceElectromyographyPhysiology2016}. Despite its promise for applications in prosthetics, rehabilitation and clinical assessment, the complexity, noise and heterogeneity of sEMG signals has limited progress in practice.

Machine learning has emerged as a promising approach to decode salient features from complex sEMG signals, achieving strong performance on tasks such as gesture classification \cite{heMultiscaleDualstreamFusion2026} \cite{sultanaSystematicReviewSurface2023} and pose regression \cite{salterEmg2poseLargeDiverse2024}. However, generalisation across sessions and subjects remains challenging due to distribution shifts arising from electrode displacement and inter-subject anatomical variability. To improve generalisation and prevent overfitting, supervised machine learning models require large labelled datasets spanning a diverse set of subjects, but collecting such datasets is infeasible for many tasks, especially in clinical applications.

Foundation models address data scarcity through a self-supervised pretraining phase on large unlabelled datasets, during which the model learns general-purpose representations that capture the underlying structure of biosignal data across diverse subjects and recording conditions. These representations can then be transferred to data-limited downstream tasks through supervised fine-tuning, reducing reliance on large labelled datasets.

A small number of sEMG foundation models have recently emerged, each adopting the patching and masked modelling approach established in the EEG foundation model literature. TinyMyo \cite{fasuloTinyMyoTinyFoundation2026} is a lightweight transformer pretrained to reconstruct masked patches of sEMG via a smooth L1 loss. PhysioWave \cite{chenPhysioWaveMultiScaleWaveletTransformer2025} applies a learnable multi-scale wavelet decomposition before patching and masked reconstruction of wavelet features via a transformer encoder and MLP decoder. NeuroRVQ \cite{barmpasNeuroRVQMultiScaleBiosignal2026} is a codebook-based model that predicts discrete token indices generated by a pretrained VQ-VAE tokeniser, reframing pretraining as a classification problem over a fixed vocabulary. However, the evaluation strategies employed by these models have several limitations that make it difficult to determine whether strong downstream performance reflects genuine knowledge transfer from pretraining, or simply larger model capacity.

First, most models are evaluated using intra-session splits, training and testing on different repetitions from the same recording session. This fails to evaluate generalisation to unseen subjects and electrode configurations, which is the setting most relevant for real-world applications. Second, existing works often do not ablate the effect of the pretraining phase directly. Without a from-scratch baseline, it is impossible to determine whether performance gains are due to pretraining or simply increased model capacity. Finally, inconsistencies in preprocessing, prediction heads and data pipelines across models mean that even where results exist, direct comparison is confounded by factors unrelated to the foundation model itself.

To address these limitations, we conduct a systematic evaluation of three sEMG foundation models under both intra-session and inter-subject splits using three finetuning strategies which isolate the contribution of pretraining from model capacity. Our findings reveal a gap between intra-session and inter-subject generalisation: pretraining provides a clear performance boost when models are fine-tuned and evaluated within the same session, but this benefit largely disappears under inter-subject evaluation. In low-data inter-subject regimes, classical ML baselines with handcrafted features match or exceed foundation model performance, suggesting that current sEMG foundation models do not yet alleviate the cross-subject overfitting that makes supervised sEMG learning so challenging.

\section{Methodology}

\subsection{Datasets \& Splits}

For each dataset, we evaluate using two split strategies which determine the level of generalisation being evaluated: an \textbf{Intra-Session} split where training and test samples are drawn from the same session and subject, and an \textbf{Inter-Subject} split where training and test sets are drawn from entirely different subjects, evaluating generalisation to unseen individuals. We split each dataset into train, validation and test partitions with a ratio of approximately 7:1:2. 

\paragraph{NinaPro DB5} NinaPro DB5 \cite{pizzolatoComparisonSixElectromyography2017} is a simultaneous sEMG and kinematics dataset in which subjects were prompted to perform 6 repetitions of 52 gestures, grouped into three exercises with rest periods between gestures. sEMG was recorded at 200\,Hz using two 8-channel Thalmic Myo armbands placed adjacently on the forearm, with the distal band rotated 22.5° relative to the proximal band. The dataset comprises 10 subjects, and recordings were segmented by gesture and divided into 1\,s windows with a stride of 250\,ms for downstream evaluation.


\paragraph{EPN612} EPN612 \cite{benalcazarEMGEPN612Dataset2020} is a large-scale gesture classification dataset comprising sEMG recordings from 612 subjects performing 6 gestures, including rest. sEMG was recorded at 200\,Hz using a single Thalmic Myo armband placed on the forearm, with each subject performing 50 repetitions per gesture within approximately 5\,s recording windows. As gesture onset labels are unreliable, we segment a single 1\,s window per recording by locating the peak of a smoothed RMS envelope using a sliding window, yielding one window per repetition for downstream evaluation.

\paragraph{UCI EMG Gestures} UCI EMG Gestures \cite{n.krilovaEMGDataGestures2018} is a gesture classification dataset with sEMG recorded at 200\,Hz using an 8-channel Thalmic Myo armband on the forearm. The dataset nominally contains 7 gesture classes, however gesture 7 was not performed by all subjects and is excluded, leaving 6 classes across 36 subjects. Raw recordings were oversampled during data collection and must be corrected prior to use. We remove duplicate samples and interpolate back to a uniform 200\,Hz grid using nearest-neighbour label interpolation and linear signal interpolation. Recordings were segmented by gesture and divided into 1\,s windows with a stride of 250\,ms.

To improve generalisation and reduce overfitting, two data augmentations were applied: a random amplitude scale, with scale factor uniformly sampled between 0.9 and 1.1, and a random channel dropout with 5\% dropout probability to simulate dead channels. A random temporal jitter is also applied to the windowing process to introduce variability in the start time of each training window. 

\subsection{Preprocessing}
\label{sec:preproc}

During our evaluation, we identified discrepancies between the preprocessing pipelines described in published papers and their released implementations. In some cases, we were unable to replicate reported results using the preprocessing steps outlined in the original papers. We therefore conducted a systematic ablation to determine the most effective normalisation and resampling strategy for each model (Appendix \ref{app:preproc}), ensuring that downstream evaluation reflects each model's true capability rather than suboptimal preprocessing choices. 

The final preprocessing pipeline for each model is shown in Table \ref{tab:preprocessing}.

\begin{table}[ht]
\centering
\caption{Final preprocessing pipeline for each model used in all 
subsequent experiments.}
\label{tab:preprocessing}
\begin{tabular}{llll}
\toprule
\textbf{Model} & \textbf{Filtering} & \textbf{Target Fs} & 
\textbf{Normalisation} \\
\midrule
NeuroRVQ   & 20--90\,Hz BPF        & 1000\,Hz & Min-max $[-1,1]$ \\
PhysioWave & None                  & 2000\,Hz & Min-max $[-1,1]$ \\
TinyMyo    & 20--90\,Hz BPF + notch & 2000\,Hz & Global $z$-score \\
\bottomrule
\end{tabular}
\end{table}

\subsection{Fine-tuning Protocol}
\label{sec:finetune}

All models are evaluated using a standardised fine-tuning pipeline. Preprocessed sEMG windows are passed through the foundation model encoder to produce per-patch embeddings, which are aggregated using mean pooling to yield a single embedding per window. A prediction head then projects this embedding to the target space. We use a single linear layer as the prediction head as a more expressive head could compensate for weak representations. The loss is computed using cross-entropy for classification, and backpropagated through the unfrozen portions of the network.

The fine-tuning strategy defines how the backbone is initialised and whether its weights are updated during fine-tuning. The three strategies we evaluate are summarised in Table~\ref{tab:finetuning_strategies}, each designed to ablate a different aspect of the foundation model hypothesis. Full fine-tuning uses the pretrained backbone and trains the full model end-to-end. From-scratch also trains end-to-end, but with a randomly initialised backbone. If these two strategies achieve similar performance, then the model architecture may be effective for supervised learning, but the pretrained representations have not transferred. Linear probing evaluates the linear separability of pretrained representations with the backbone frozen. A model that performs well here has learned features that are directly useful for gesture classification without any task-specific adaptation.


\begin{table}[ht]
\centering
\caption{The three fine-tuning strategies used for evaluation, 
defined by backbone initialisation and whether backbone weights 
are updated during training.}
\label{tab:finetuning_strategies}
\begin{tabular}{@{}lcc@{}}
\toprule
\textbf{Strategy} & \textbf{Backbone Init} & \textbf{Backbone Weights} \\
\midrule
Full Fine-tuning  & Pretrained & Updated \\
From-Scratch      & Random     & Updated \\
Linear Probing    & Pretrained & Frozen  \\
\bottomrule
\end{tabular}
\end{table}

All experiments were trained using AdamW with a weight decay of 0.001 (excluding biases, normalisation layers, and spatial/positional embeddings), with a 5-epoch linear learning rate warmup followed by cosine annealing decay. An initial hyperparameter search selected the most effective learning rate for each model at a batch size of 32. For each subsequent experiment, the maximum batch size fitting in memory was found automatically, with gradient accumulation used when this fell below 32. The base learning rate was then scaled linearly with batch size following \cite{goyalAccurateLargeMinibatch2018}. All models were trained to convergence via early stopping with a patience of 8 epochs. Each experiment was repeated with three random seeds and standard deviations are reported as error bounds.

To contextualize foundation model performance, we evaluated classical ML algorithms (linear discriminant analysis, support vector machine and random forest) trained on the Hudgins time-domain feature set~\cite{hudginsNewStrategyMultifunction1993}, comprising mean absolute value, waveform length, zero-crossings and slope sign changes computed per channel. We also include a simple supervised CNN model with two 1D convolution layers with a kernel size of 3, and two fully-connected layers.

\section{Results}

\begin{table}[]
\centering
\resizebox{\columnwidth}{!}{%
\begin{tabular}{llcccccc}
\toprule
 & & \multicolumn{2}{c}{\textbf{EPN-612 (1.0s)}} & \multicolumn{2}{c}{\textbf{NinaPro DB5 (1.0s)}} & \multicolumn{2}{c}{\textbf{UCI-EMG (1.0s)}} \\
\cmidrule(lr){3-4}\cmidrule(lr){5-6}\cmidrule(lr){7-8}
Model & Tuning & Acc (\%) & F1 Score & Acc (\%) & F1 Score & Acc (\%) & F1 Score \\
\midrule
LDA & - & 75.33 $\pm$ 0.03 & 75.73 $\pm$ 0.03 & 30.84 $\pm$ 0.10 & 30.39 $\pm$ 0.13 & 84.40 $\pm$ 0.00 & 84.47 $\pm$ 0.01 \\
SVM & - & 87.06 $\pm$ 0.04 & 87.08 $\pm$ 0.04 & 65.35 $\pm$ 0.18 & 65.22 $\pm$ 0.24 & 91.89 $\pm$ 0.54 & 91.85 $\pm$ 0.55 \\
RF & - & 87.25 $\pm$ 0.06 & 87.26 $\pm$ 0.06 & 71.36 $\pm$ 0.54 & 71.22 $\pm$ 0.53 & 89.76 $\pm$ 0.07 & 89.66 $\pm$ 0.06 \\
\midrule
CNN & - & 94.89 $\pm$ 0.10 & 94.89 $\pm$ 0.10 & 41.06 $\pm$ 1.02 & 39.70 $\pm$ 0.92 & 81.65 $\pm$ 0.62 & 81.52 $\pm$ 0.62 \\
\midrule
NeuroRVQ & Full FT & \textbf{97.48 $\pm$ 0.07} & \textbf{97.48 $\pm$ 0.07} & \textbf{\underline{84.15 $\pm$ 0.46}} & \textbf{\underline{84.02 $\pm$ 0.54}} & \textbf{\underline{94.10 $\pm$ 1.21}} & \textbf{\underline{94.11 $\pm$ 1.19}} \\
NeuroRVQ & From-Scratch & 93.84 $\pm$ 0.27 & 93.85 $\pm$ 0.26 & 76.21 $\pm$ 0.58 & 75.92 $\pm$ 0.61 & 87.18 $\pm$ 0.47 & 87.18 $\pm$ 0.53 \\
NeuroRVQ & LP & 90.57 $\pm$ 0.20 & 90.59 $\pm$ 0.20 & 49.28 $\pm$ 0.81 & 48.30 $\pm$ 0.94 & 90.29 $\pm$ 0.89 & 90.32 $\pm$ 0.87 \\
\midrule
PhysioWave & Full FT & \textbf{\underline{97.63 $\pm$ 0.05}} & \textbf{\underline{97.62 $\pm$ 0.05}} & 72.14 $\pm$ 0.45 & 71.84 $\pm$ 0.50 & \textbf{92.59 $\pm$ 0.47} & \textbf{92.63 $\pm$ 0.46} \\
PhysioWave & From-Scratch & 97.14 $\pm$ 0.11 & 97.15 $\pm$ 0.12 & \textbf{72.68 $\pm$ 0.84} & \textbf{72.43 $\pm$ 0.80} & 88.41 $\pm$ 2.76 & 88.46 $\pm$ 2.72 \\
PhysioWave & LP & 86.60 $\pm$ 0.06 & 86.64 $\pm$ 0.06 & 23.09 $\pm$ 0.07 & 21.10 $\pm$ 0.02 & 84.28 $\pm$ 0.25 & 84.20 $\pm$ 0.27 \\
\midrule
TinyMyo & Full FT & \textbf{64.40 $\pm$ 0.67} & \textbf{64.40 $\pm$ 0.80} & \textbf{16.43 $\pm$ 1.26} & \textbf{15.18 $\pm$ 1.35} & \textbf{40.75 $\pm$ 1.35} & \textbf{39.34 $\pm$ 1.81} \\
TinyMyo & From-Scratch & 63.00 $\pm$ 0.56 & 62.98 $\pm$ 0.76 & 15.62 $\pm$ 0.39 & 14.76 $\pm$ 0.42 & 39.76 $\pm$ 1.58 & 38.88 $\pm$ 0.92 \\
TinyMyo & LP & 41.91 $\pm$ 0.20 & 39.06 $\pm$ 0.38 & 4.21 $\pm$ 0.18 & 1.08 $\pm$ 0.21 & 26.66 $\pm$ 2.81 & 19.64 $\pm$ 3.61 \\
\bottomrule
\end{tabular}
}
\caption{Intra-session gesture classification accuracy and 
F1 score. \textbf{Bold} 
indicates the best result per model per dataset; \textbf{\underline{bold 
underline}} indicates the best overall per dataset.}
\label{tab:intra_session}
\end{table}

\begin{table}[]
\centering
\resizebox{\columnwidth}{!}{%
\begin{tabular}{llcccccc}
\toprule
 &  & \multicolumn{2}{c}{\textbf{EPN-612 (1.0s)}} & \multicolumn{2}{c}{\textbf{NinaPro DB5 (1.0s)}} & \multicolumn{2}{c}{\textbf{UCI-EMG (1.0s)}} \\
\cmidrule(lr){3-4}\cmidrule(lr){5-6}\cmidrule(lr){7-8}
Model & Tuning & Acc (\%) & F1 Score & Acc (\%) & F1 Score & Acc (\%) & F1 Score \\
\midrule
LDA & - & 74.91 $\pm$ 0.07 & 75.38 $\pm$ 0.06 & \textbf{\underline{23.63 $\pm$ 0.09}} & \textbf{\underline{21.97 $\pm$ 0.13}} & 88.32 $\pm$ 0.28 & 88.36 $\pm$ 0.28 \\
SVM & - & 81.69 $\pm$ 0.09 & 81.71 $\pm$ 0.10 & 23.43 $\pm$ 0.51 & 21.34 $\pm$ 0.67 & \textbf{\underline{89.41 $\pm$ 0.46}} & \textbf{\underline{89.38 $\pm$ 0.47}} \\
RF & - & 80.20 $\pm$ 0.03 & 80.26 $\pm$ 0.03 & 22.15 $\pm$ 0.58 & 20.46 $\pm$ 0.60 & 89.17 $\pm$ 1.03 & 89.20 $\pm$ 1.04 \\
\midrule
CNN & - & 91.48 $\pm$ 0.44 & 91.50 $\pm$ 0.45 & 22.59 $\pm$ 0.41 & 18.68 $\pm$ 0.03 & 77.01 $\pm$ 1.05 & 76.28 $\pm$ 1.01 \\
\midrule
NeuroRVQ & Full FT & \textbf{\underline{93.26 $\pm$ 0.27}} & \textbf{\underline{93.30 $\pm$ 0.26}} & 23.58 $\pm$ 1.15 & 20.51 $\pm$ 1.92 & 83.91 $\pm$ 1.63 & 83.49 $\pm$ 2.03 \\
NeuroRVQ & From-Scratch & 90.94 $\pm$ 0.32 & 90.94 $\pm$ 0.29 & \textbf{23.62 $\pm$ 0.04} & \textbf{20.68 $\pm$ 0.27} & 81.13 $\pm$ 3.77 & 81.09 $\pm$ 4.20 \\
NeuroRVQ & LP & 88.61 $\pm$ 0.36 & 88.62 $\pm$ 0.42 & 23.55 $\pm$ 0.99 & 20.49 $\pm$ 0.99 & \textbf{88.63 $\pm$ 0.64} & \textbf{88.69 $\pm$ 0.65} \\
\midrule
PhysioWave & Full FT & \textbf{92.46 $\pm$ 0.42} & \textbf{92.46 $\pm$ 0.40} & 19.03 $\pm$ 1.25 & 16.73 $\pm$ 1.72 & 83.73 $\pm$ 1.82 & 83.48 $\pm$ 1.82 \\
PhysioWave & From-Scratch & 92.31 $\pm$ 0.47 & 92.32 $\pm$ 0.46 & \textbf{23.25 $\pm$ 1.09} & \textbf{21.51 $\pm$ 0.86} & 82.58 $\pm$ 1.10 & 82.60 $\pm$ 1.09 \\
PhysioWave & LP & 86.19 $\pm$ 0.09 & 86.25 $\pm$ 0.09 & 14.05 $\pm$ 0.22 & 13.28 $\pm$ 0.34 & \textbf{85.18 $\pm$ 0.46} & \textbf{85.08 $\pm$ 0.44} \\
\midrule
TinyMyo & Full FT & \textbf{62.63 $\pm$ 0.55} & \textbf{62.51 $\pm$ 0.15} & \textbf{8.62 $\pm$ 0.33} & \textbf{6.45 $\pm$ 0.18} & 38.05 $\pm$ 1.41 & 35.73 $\pm$ 1.59 \\
TinyMyo & From-Scratch & 62.59 $\pm$ 0.86 & 62.31 $\pm$ 0.88 & 7.77 $\pm$ 0.87 & 5.38 $\pm$ 0.55 & \textbf{40.65 $\pm$ 1.61} & \textbf{37.69 $\pm$ 1.21} \\
TinyMyo & LP & 40.52 $\pm$ 0.10 & 35.95 $\pm$ 0.53 & 2.55 $\pm$ 0.40 & 0.25 $\pm$ 0.14 & 27.47 $\pm$ 1.06 & 21.73 $\pm$ 2.73 \\
\bottomrule
\end{tabular}
}
\caption{Inter-subject gesture classification accuracy and F1 score.
\textbf{Bold} indicates the best result per model per dataset; 
\textbf{\underline{bold underline}} indicates the best overall 
per dataset.}
\label{tab:inter_subject}
\end{table}

Tables~\ref{tab:intra_session} and~\ref{tab:inter_subject} present gesture classification accuracy and F1 score for all models and fine-tuning strategies on the three datasets. 

Table~\ref{tab:intra_session} presents intra-session results, where models are finetuned and evaluated on different repetitions from the same session. These results broadly support the foundation model claim: pretrained initialisation consistently outperforms random initialisation across models and datasets. The exception is PhysioWave on NinaPro DB5, where From Scratch outperforms Full Fine-tuning, suggesting pretrained representations have failed to transfer. NeuroRVQ shows the strongest pretraining benefit and achieves the highest accuracy in low-data regimes (NinaPro DB5 and UCI-EMG), while PhysioWave performs best on the larger EPN-612 dataset. NeuroRVQ's strong linear probing performance provides the clearest evidence of genuine representation quality among the evaluated models. TinyMyo performs substantially below the other two models across all datasets, which may be due its smaller embedding dimensionality and resulting linear head. We ablate the pooling strategy in Appendix~\ref{app:pooling} and find that results improve with a pooling strategy matching the original paper, though TinyMyo remains the weakest model overall. 

These results contrast sharply with Table~\ref{tab:inter_subject}, where models were finetuned and tested on entirely different subjects. On EPN-612, we still see the same expected performance increases across tuning strategies, with full fine-tuning outperforming training from scratch, although to a lesser extent. However, on NinaPro DB5 and UCI-EMG datasets, all models show little to no pretraining benefit, and classical ML baselines match or exceed foundation model performance. Despite showing clear pretraining benefit within sessions, current foundation models alone fail to overcome the cross-subject overfitting that makes supervised sEMG learning so challenging in low-data regimes.

Our results differ from those reported in the original papers for several reasons. First, we use 1\,s windows throughout, whereas NeuroRVQ reports results with 5\,s windows on EPN-612 and NinaPro DB5. Second, our primary evaluation uses inter-subject splits, whereas inspection of the released implementations of TinyMyo and PhysioWave reveals that their published results use intra-session splits. Third, TinyMyo's includes rest periods between gestures as an additional class on NinaPro DB5, introducing substantial class imbalance and label leakage from zero-padded windows. Finally, NeuroRVQ and TinyMyo use pooling strategies involving partial concatenation of patch embeddings, whereas our pipeline uses mean pooling throughout. We ablate pooling strategy in Appendix \ref{app:pooling}.

\section{Discussion and Conclusion}


Taken together, these results show that pretraining alone is not sufficient to prevent cross-subject overfitting on data-limited downstream tasks. This finding is not entirely unexpected, as reconstruction-based pretraining objectives do not incentivise the model to discard subject-specific features, and therefore models are still likely to overfit downstream datasets when the number of training subjects is small. While retaining subject-specific features may be beneficial for some tasks — such as predicting clinical biomarkers — for gesture classification and pose regression, learning subject-invariant representations would help reduce overfitting. 

These findings point to several directions for future work. First, new fine-tuning strategies are needed that can improve cross-subject knowledge transfer while limiting overfitting in low-data regimes. Second, the effect of varying the number of training subjects available for fine-tuning should be explored, as datasets like NinaPro DB5 may simply not contain enough subjects for effective inter-subject generalisation. Finally, future work should explore alternative pretraining objectives and architectures to improve knowledge transfer, as even on EPN-612 — the largest downstream dataset evaluated — pretraining provides a maximum accuracy gain of just 2.32\%. More broadly, future evaluations of sEMG foundation models should prioritise inter-subject generalisation as the primary evaluation setting, and should always include a from-scratch baseline to isolate the contribution of the pretraining phase.

\newpage
\bibliographystyle{unsrt} 
\bibliography{references}


\appendix

\section{Preprocessing Ablations}
\label{app:preproc}

We conducted a systematic ablation to determine the most effective normalisation and resampling strategy for each model, ensuring that downstream evaluation reflects each model's true capability rather than suboptimal preprocessing choices.

Ablation experiments use LOSO cross-validation on the training and validation subjects of the NinaPro DB5 and UCI Gestures datasets to prevent test set contamination. Each ablation varies only the preprocessing stage, with all other pipeline stages held fixed. Ablations are evaluated under both the head-only and full fine-tuning protocols. Filtering is held fixed across all experiments and kept consistent with each model's original implementation, as summarised in Table~\ref{tab:preprocessing}.

\begin{figure}
    \centering
    \includegraphics[width=\linewidth]{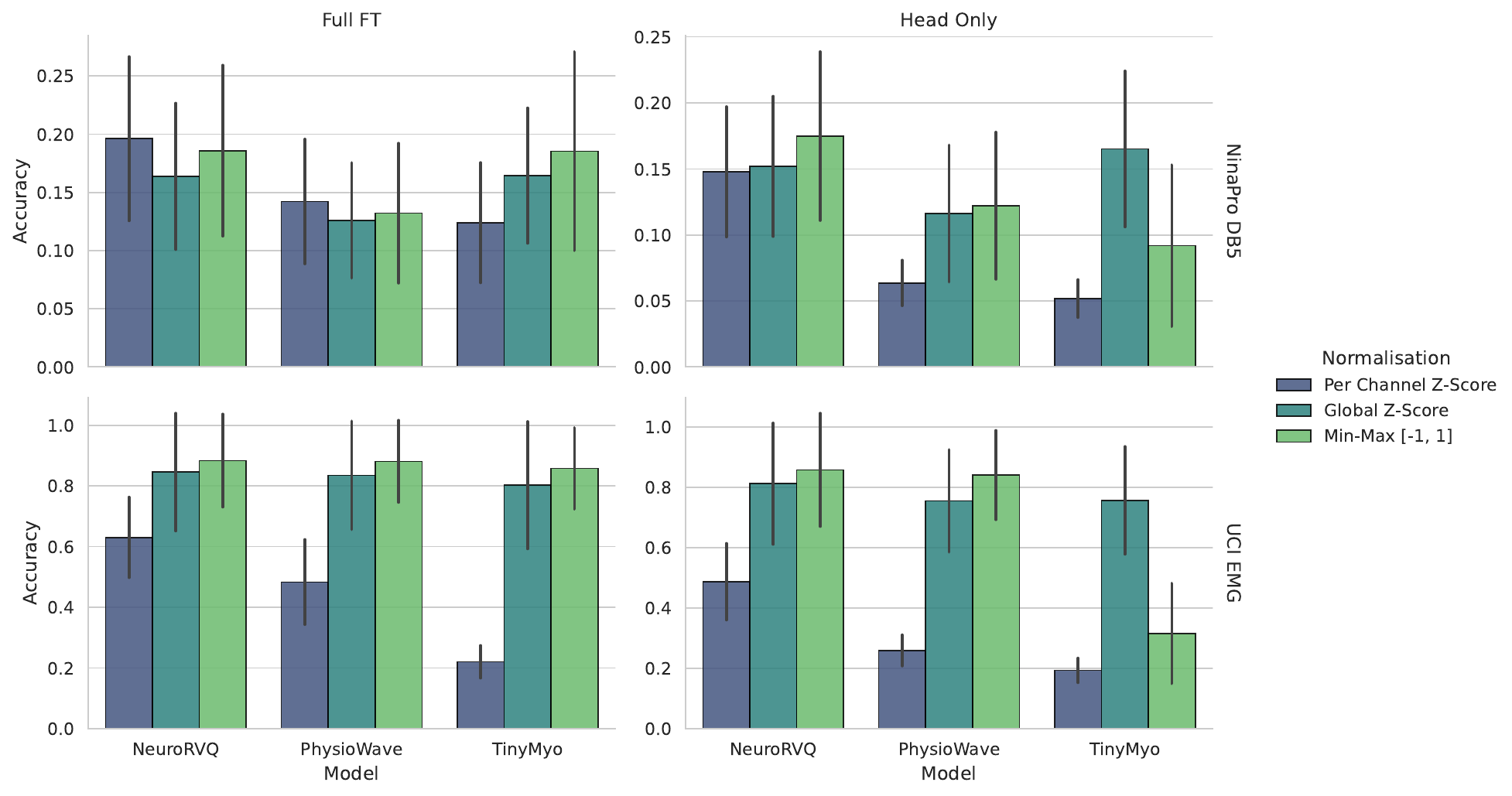}
    \caption{Gesture classification accuracy under three normalisation strategies (per-channel z-score, global z-score, min-max $[-1,1]$) for each model on NinaPro DB5 and UCI-EMG, evaluated under full fine-tuning and head-only protocols.}
    \label{fig:norm}
\end{figure}

We first ablate normalisation strategy (Figure \ref{fig:norm}), evaluating three approaches: per-channel $z$-score, global $z$-score, and min-max normalisation to $[-1, 1]$ (formulations given in Table~\ref{tab:normalisation_equations}). To isolate the effect of normalisation, all signals are resampled to each model's pretraining frequency via sinc interpolation with a Hann window throughout this ablation. We selected the strategy that maximised head-only performance for each model, as this is the most sensitive condition: min-max normalisation for PhysioWave and NeuroRVQ, and global $z$-score for TinyMyo. 

\begin{figure}
    \centering
    \includegraphics[width=\linewidth]{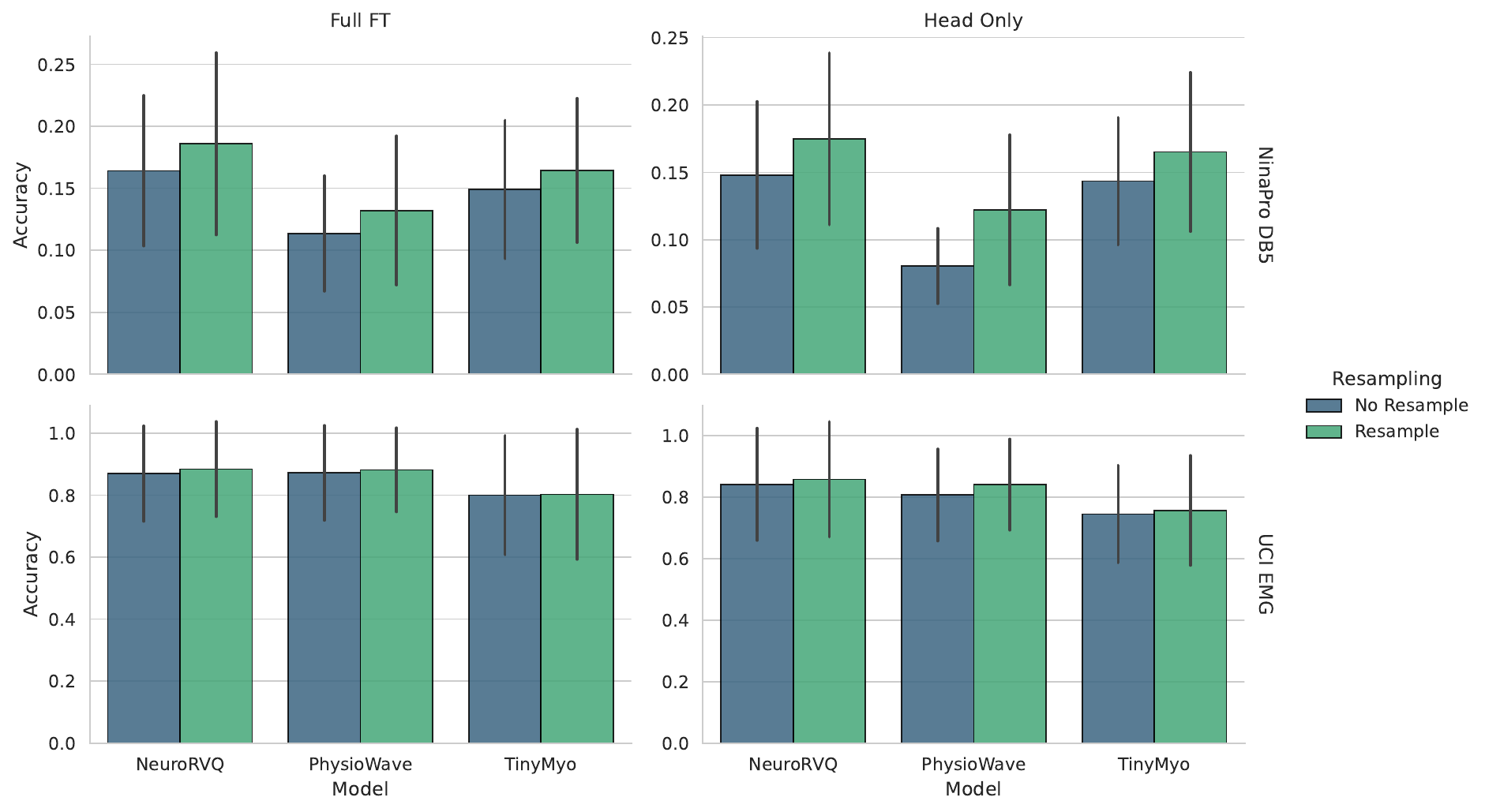}
    \caption{Effect of resampling on gesture classification accuracy. Each model is evaluated with and without upsampling to its pretraining frequency, under full fine-tuning and head-only protocols on NinaPro DB5 and UCI-EMG.}
    \label{fig:resample}
\end{figure}

With these normalisation strategies held fixed, we evaluated the effect of resampling on gesture classification performance. Two conditions were compared: a no-resample baseline, which leaves the signal at its original sampling frequency (200\,Hz), and a resample condition, which uses band-limited sinc interpolation with a Hann window to upsample each signal to match the pretraining frequency of each model — 1000\,Hz for NeuroRVQ and 2000\,Hz for PhysioWave and TinyMyo. The results are presented in Figure~\ref{fig:resample}. Resampling improved performance across every model, dataset, and fine-tuning combination tested, confirming that matching the input frequency distribution to pretraining conditions is important.

\begin{table}[ht]
\centering
\caption{Mathematical formulations of the three normalisation strategies, 
where $\tilde{X}$ denotes the normalised signal and $\epsilon$ is a small 
constant for numerical stability.}
\label{tab:normalisation_equations}
\begin{tabular}{lll}
\toprule
\textbf{Strategy} & \textbf{Formulation} & \textbf{Statistics Computed Over} \\
\midrule
Per-channel $z$-score 
    & $\tilde{X}_{c,t} = \dfrac{X_{c,t} - \mu_c}{\sigma_c + \epsilon}$ 
    & $\mu_c,\, \sigma_c$ over $T$ (per channel) \\[12pt]
Global $z$-score 
    & $\tilde{X}_{c,t} = \dfrac{X_{c,t} - \mu}{\sigma + \epsilon}$ 
    & $\mu,\, \sigma$ over $C \times T$ \\[12pt]
Min-max $[-1, 1]$ 
    & $\tilde{X}_{c,t} = 2\!\left(\dfrac{X_{c,t} - X_{\min}}
      {X_{\max} - X_{\min} + \epsilon}\right) - 1$ 
    & $X_{\min},\, X_{\max}$ over $C \times T$ \\
\bottomrule
\end{tabular}
\end{table}

\section{Pooling Ablation}
\label{app:pooling}

Tables~\ref{tab:concat_pooling_subject} and \ref{tab:concat_pooling_session} present gesture classification results using channel-preserving mean pooling, which concatenates patch embeddings across the channel dimension before mean pooling across time. This matches the pooling strategy used in the original TinyMyo and NeuroRVQ implementations, and is provided to contextualise the lower performance of these models under our standardised mean pooling pipeline. TinyMyo in particular improves substantially — from 62.63\% to 83.95\% on EPN-612 under full fine-tuning — confirming that its underperformance in the main results partly reflects a pipeline mismatch rather than model quality alone.

\begin{table}[]
\centering
\resizebox{\columnwidth}{!}{%
\begin{tabular}{llcccccc}
\toprule
 &  & \multicolumn{2}{c}{\textbf{EPN-612 (1s)}} & \multicolumn{2}{c}{\textbf{NinaPro DB5 (1s)}} & \multicolumn{2}{c}{\textbf{UCI-EMG (1s)}} \\
\cmidrule(lr){3-4}\cmidrule(lr){5-6}\cmidrule(lr){7-8}
Model & Tuning & Acc (\%) & F1 Score & Acc (\%) & F1 Score & Acc (\%) & F1 Score \\
\midrule
NeuroRVQ & Full FT & \textbf{\underline{92.75 $\pm$ 0.57}} & \textbf{\underline{92.75 $\pm$ 0.60}} & \textbf{\underline{26.92 $\pm$ 0.86}} & \textbf{\underline{23.91 $\pm$ 1.70}} & 88.26 $\pm$ 0.52 & 88.18 $\pm$ 0.61 \\
NeuroRVQ & From-Scratch FT & 90.55 $\pm$ 0.24 & 90.55 $\pm$ 0.26 & 24.11 $\pm$ 1.91 & 20.53 $\pm$ 1.64 & 87.11 $\pm$ 1.84 & 86.97 $\pm$ 1.93 \\
NeuroRVQ & LP & 90.38 $\pm$ 0.04 & 90.41 $\pm$ 0.06 & 25.54 $\pm$ 0.68 & 21.91 $\pm$ 0.81 & \textbf{\underline{88.45 $\pm$ 1.06}} & \textbf{\underline{88.45 $\pm$ 1.05}} \\
\midrule
PhysioWave & Full FT & 92.24 $\pm$ 0.63 & 92.27 $\pm$ 0.66 & \textbf{20.00 $\pm$ 2.11} & \textbf{17.66 $\pm$ 2.37} & \textbf{85.66 $\pm$ 1.96} & \textbf{85.54 $\pm$ 2.02} \\
PhysioWave & From-Scratch FT & \textbf{92.38 $\pm$ 0.28} & \textbf{92.38 $\pm$ 0.28} & 19.37 $\pm$ 1.47 & 17.05 $\pm$ 1.01 & 84.03 $\pm$ 1.26 & 83.81 $\pm$ 1.33 \\
PhysioWave & LP & 89.36 $\pm$ 0.16 & 89.43 $\pm$ 0.16 & 18.07 $\pm$ 0.91 & 16.71 $\pm$ 1.42 & 84.21 $\pm$ 1.57 & 84.01 $\pm$ 1.58 \\
\midrule
TinyMyo & Full FT & \textbf{83.95 $\pm$ 0.35} & \textbf{83.92 $\pm$ 0.36} & \textbf{26.00 $\pm$ 0.72} & \textbf{21.41 $\pm$ 0.94} & 80.22 $\pm$ 3.93 & 79.77 $\pm$ 3.83 \\
TinyMyo & From-Scratch FT & 83.29 $\pm$ 0.03 & 83.35 $\pm$ 0.04 & 24.23 $\pm$ 1.16 & 19.88 $\pm$ 1.03 & 81.91 $\pm$ 3.12 & 81.52 $\pm$ 3.48 \\
TinyMyo & LP & 75.28 $\pm$ 0.07 & 75.01 $\pm$ 0.09 & 21.60 $\pm$ 0.51 & 16.61 $\pm$ 0.49 & 83.67 $\pm$ 0.83 & 83.48 $\pm$ 0.93 \\
\bottomrule
\end{tabular}
}
\caption{Inter-subject gesture classification results using channel-concat 
pooling (patch embeddings concatenated across channels, mean pooled 
across time), matching the pooling strategy used in the original 
TinyMyo and NeuroRVQ implementations. Compare with 
Table~\ref{tab:inter_subject} to isolate the effect of pooling 
strategy.}
\label{tab:concat_pooling_subject}
\end{table}

\begin{table}[]
\centering
\resizebox{\columnwidth}{!}{%
\begin{tabular}{llcccccc}
\toprule
 &  & \multicolumn{2}{c}{\textbf{EPN-612 (1.0s)}} & \multicolumn{2}{c}{\textbf{NinaPro DB5 (1.0s)}} & \multicolumn{2}{c}{\textbf{UCI-EMG (1.0s)}} \\
\cmidrule(lr){3-4}\cmidrule(lr){5-6}\cmidrule(lr){7-8}
Model & Tuning & Acc (\%) & F1 Score & Acc (\%) & F1 Score & Acc (\%) & F1 Score \\
\midrule
NeuroRVQ & Full FT & \textbf{97.45 $\pm$ 0.05} & \textbf{97.46 $\pm$ 0.06} & \textbf{\underline{84.21 $\pm$ 0.74}} & \textbf{\underline{84.11 $\pm$ 0.77}} & \textbf{93.78 $\pm$ 1.36} & \textbf{93.82 $\pm$ 1.36} \\
NeuroRVQ & From-Scratch & 93.61 $\pm$ 0.35 & 93.62 $\pm$ 0.35 & 73.41 $\pm$ 1.07 & 73.17 $\pm$ 1.03 & 91.24 $\pm$ 0.43 & 91.29 $\pm$ 0.43 \\
NeuroRVQ & LP & 92.94 $\pm$ 0.26 & 92.94 $\pm$ 0.26 & 74.37 $\pm$ 0.39 & 74.26 $\pm$ 0.46 & 92.67 $\pm$ 0.79 & 92.65 $\pm$ 0.79 \\
\midrule
PhysioWave & Full FT & \textbf{\underline{97.64 $\pm$ 0.15}} & \textbf{\underline{97.64 $\pm$ 0.15}} & \textbf{75.40 $\pm$ 1.85} & \textbf{75.14 $\pm$ 1.97} & \textbf{\underline{94.64 $\pm$ 0.68}} & \textbf{\underline{94.70 $\pm$ 0.66}} \\
PhysioWave & From-Scratch & 97.21 $\pm$ 0.19 & 97.22 $\pm$ 0.19 & 72.92 $\pm$ 1.78 & 72.71 $\pm$ 1.70 & 91.24 $\pm$ 1.02 & 91.17 $\pm$ 1.03 \\
PhysioWave & LP & 91.69 $\pm$ 0.47 & 91.73 $\pm$ 0.47 & 41.42 $\pm$ 0.10 & 40.52 $\pm$ 0.13 & 91.40 $\pm$ 0.65 & 91.34 $\pm$ 0.67 \\
\midrule
TinyMyo & Full FT & \textbf{89.55 $\pm$ 0.34} & \textbf{89.55 $\pm$ 0.34} & \textbf{64.77 $\pm$ 0.14} & \textbf{64.27 $\pm$ 0.16} & \textbf{85.22 $\pm$ 0.91} & \textbf{85.30 $\pm$ 0.93} \\
TinyMyo & From-Scratch & 88.92 $\pm$ 0.25 & 88.93 $\pm$ 0.26 & 63.84 $\pm$ 0.81 & 63.39 $\pm$ 0.81 & 84.85 $\pm$ 0.86 & 84.91 $\pm$ 0.89 \\
TinyMyo & LP & 76.74 $\pm$ 0.06 & 76.73 $\pm$ 0.10 & 32.76 $\pm$ 0.09 & 30.73 $\pm$ 0.16 & 77.97 $\pm$ 0.62 & 77.93 $\pm$ 0.62 \\
\bottomrule
\end{tabular}
}
\caption{Intra-session gesture classification results using channel-concat 
pooling (patch embeddings concatenated across channels, mean pooled 
across time), matching the pooling strategy used in the original 
TinyMyo and NeuroRVQ implementations. Compare with 
Table~\ref{tab:intra_session} to isolate the effect of pooling 
strategy.}
\label{tab:concat_pooling_session}
\end{table}

\section{Original Preprocessing Pipelines}
\label{app:original_pipelines}

To address reviewer feedback regarding model-specific preprocessing, we also evaluated each model under its original preprocessing pipeline as described in the papers (Tables~\ref{tab:fft_original_preproc_results} \&~\ref{tab:lp_original_preproc_results}). Although the original preprocessing pipelines did provide a small improvement in some conditions (PhysioWave Full FT on EPN-612 and NinaPro DB5), the optimised preprocessing pipeline outperforms the native pipeline under most conditions. The effect is especially prevalent on the UCI-EMG dataset and in the linear probing setting, where the optimised preprocessing pipeline provides a massive increase in performance. This is likely due to these settings having a greater reliance on relative amplitude information between channels, which is destroyed by the native per-channel z-score normalisation. 

\begin{table}[ht]
\centering
\resizebox{\columnwidth}{!}{%
\begin{tabular}{llcccccc}
\toprule
 &  & \multicolumn{2}{c}{\textbf{EPN-612 (1.0s)}} & \multicolumn{2}{c}{\textbf{NinaPro DB5 (1.0s)}} & \multicolumn{2}{c}{\textbf{UCI-EMG (1.0s)}} \\
\cmidrule(lr){3-4}\cmidrule(lr){5-6}\cmidrule(lr){7-8}
Model & Preprocessing & Acc (\%) & F1 Score & Acc (\%) & F1 Score & Acc (\%) & F1 Score \\
\midrule
NeuroRVQ & Native & \textbf{\underline{93.26 $\pm$ 0.27}} & \textbf{\underline{93.30 $\pm$ 0.26}} & \textbf{\underline{23.58 $\pm$ 1.15}} & \textbf{\underline{20.51 $\pm$ 1.92}} & \textbf{\underline{83.91 $\pm$ 1.63}} & \textbf{\underline{83.49 $\pm$ 2.03}} \\
\midrule
PhysioWave & Native & \textbf{93.08 $\pm$ 0.22} & \textbf{93.09 $\pm$ 0.19} & \textbf{21.70 $\pm$ 0.85} & \textbf{19.70 $\pm$ 0.56} & 27.60 $\pm$ 9.95 & 21.91 $\pm$ 13.18 \\
PhysioWave & Optimised & 92.46 $\pm$ 0.42 & 92.46 $\pm$ 0.40 & 19.03 $\pm$ 1.25 & 16.73 $\pm$ 1.72 & \textbf{83.73 $\pm$ 1.82} & \textbf{83.48 $\pm$ 1.82} \\
\midrule
TinyMyo & Native & 58.77 $\pm$ 1.33 & 58.29 $\pm$ 2.00 & 7.14 $\pm$ 0.50 & 3.84 $\pm$ 1.10 & 17.31 $\pm$ 0.26 & 5.60 $\pm$ 0.66 \\
TinyMyo & Optimised & \textbf{62.63 $\pm$ 0.55} & \textbf{62.51 $\pm$ 0.15} & \textbf{8.62 $\pm$ 0.33} & \textbf{6.45 $\pm$ 0.18} & \textbf{38.05 $\pm$ 1.41} & \textbf{35.73 $\pm$ 1.59} \\
\bottomrule
\end{tabular}
}
\caption{Full Finetuning Inter-subject gesture classification performance comparing each model's native preprocessing pipeline with the optimised preprocessing determined from ablation in Appendix \ref{app:preproc} (1.0\,s windows, mean pooling).}
\label{tab:fft_original_preproc_results}
\end{table}

\begin{table}[ht]
\centering
\resizebox{\columnwidth}{!}{%
\begin{tabular}{llcccccc}
\toprule
& & \multicolumn{2}{c}{\textbf{EPN-612 (1.0s)}} & \multicolumn{2}{c}{\textbf{NinaPro DB5 (1.0s)}} & \multicolumn{2}{c}{\textbf{UCI-EMG (1.0s)}} \\
\cmidrule(lr){3-4}\cmidrule(lr){5-6}\cmidrule(lr){7-8}
Model & Preprocessing & Acc (\%) & F1 Score & Acc (\%) & F1 Score & Acc (\%) & F1 Score \\
\midrule
NeuroRVQ & Native & \textbf{\underline{88.61 $\pm$ 0.36}} & \textbf{\underline{88.62 $\pm$ 0.42}} & \textbf{\underline{23.55 $\pm$ 0.99}} & \textbf{\underline{20.49 $\pm$ 0.99}} & \textbf{\underline{88.63 $\pm$ 0.64}} & \textbf{\underline{88.69 $\pm$ 0.65}} \\
\midrule
PhysioWave & Native & 71.60 $\pm$ 0.04 & 71.54 $\pm$ 0.03 & 6.03 $\pm$ 0.16 & 3.69 $\pm$ 0.12 & 20.12 $\pm$ 1.73 & 17.43 $\pm$ 3.18 \\
PhysioWave & Optimised & \textbf{86.19 $\pm$ 0.09} & \textbf{86.25 $\pm$ 0.09} & \textbf{14.05 $\pm$ 0.22} & \textbf{13.28 $\pm$ 0.34} & 85.18 $\pm$ 0.46 & 85.08 $\pm$ 0.44 \\
\midrule
TinyMyo & Native & 35.88 $\pm$ 1.27 & 28.48 $\pm$ 1.90 & \textbf{3.91 $\pm$ 0.13} & \textbf{0.70 $\pm$ 0.12} & 16.02 $\pm$ 1.78 & 8.93 $\pm$ 1.36 \\
TinyMyo & Optimised & \textbf{40.52 $\pm$ 0.10} & \textbf{35.95 $\pm$ 0.53} & 2.55 $\pm$ 0.40 & 0.25 $\pm$ 0.14 & \textbf{27.47 $\pm$ 1.06} & \textbf{21.73 $\pm$ 2.73} \\
\bottomrule
\end{tabular}
}
\caption{Linear Probing Inter-subject gesture classification performance comparing each model's native preprocessing pipeline with the optimised preprocessing determined from ablation in Appendix \ref{app:preproc} (1.0\,s windows, mean pooling).}
\label{tab:lp_original_preproc_results}
\end{table}

\section{Leave-One-Subject-Out (LOSO) Validation on NinaPro DB5}
\label{app:loso_ninapro}

To address reviewer feedback regarding inter-subject generalisation and partition variance across limited subject cohorts, we provide full Leave-One-Subject-Out (LOSO) cross-validation results on the NinaPro DB5 dataset (Table~\ref{tab:loso_ninapro_results}).

\begin{table}[]
\centering
\resizebox{\columnwidth}{!}{%
\begin{tabular}{llcc}
\toprule
 &  & \multicolumn{2}{c}{\textbf{NinaPro DB5 (1.0s)}} \\
\cmidrule(lr){3-4}
Model & Tuning & Acc (\%) & F1 Score \\
\midrule
NeuroRVQ & Full FT & \textbf{19.37 $\pm$ 4.97} & \textbf{15.63 $\pm$ 4.58} \\
NeuroRVQ & From-Scratch & 16.82 $\pm$ 4.12 & 13.59 $\pm$ 4.24 \\
NeuroRVQ & LP & 16.30 $\pm$ 4.90 & 13.67 $\pm$ 4.45 \\
\midrule
PhysioWave & Full FT & 15.34 $\pm$ 4.85 & 12.52 $\pm$ 4.25 \\
PhysioWave & From-Scratch & \textbf{\underline{19.59 $\pm$ 4.79}} & \textbf{\underline{16.74 $\pm$ 4.93}} \\
PhysioWave & LP & 11.50 $\pm$ 2.78 & 9.24 $\pm$ 2.60 \\
\midrule
TinyMyo & Full FT & \textbf{7.33 $\pm$ 1.25} & \textbf{5.28 $\pm$ 0.96} \\
TinyMyo & From-Scratch & 5.80 $\pm$ 1.80 & 3.50 $\pm$ 1.83 \\
TinyMyo & LP & 3.40 $\pm$ 0.46 & 0.79 $\pm$ 0.26 \\
\bottomrule
\end{tabular}
}
\caption{Leave-One-Subject-Out (LOSO) gesture classification results on the NinaPro DB5 dataset using 1.0\,s windows and mean pooling.}
\label{tab:loso_ninapro_results}
\end{table}


\end{document}